\documentclass[10pt,twocolumn,letterpaper]{article}

\usepackage{iccv}
\usepackage{times}
\usepackage{epsfig}
\usepackage{graphicx}
\usepackage{amsmath}
\usepackage{amssymb}
\usepackage{booktabs}
\usepackage{bm}
\usepackage{bbm}
\usepackage{caption}
\usepackage[boxed]{algorithm2e}
\usepackage{multirow}
\usepackage{arydshln}
\usepackage[normalem]{ulem}
\usepackage{setspace}
\usepackage{xspace}
\usepackage{authblk}

\makeatletter 
\providecommand{\appendiargdef}[2]{\begingroup
\toks@\expandafter{#1{##1}#2}%
\edef\@bsx{\endgroup \def\noexpand#1####1{\the\toks@}}%
\@bsx}
\appendiargdef{\thanks}{%
\protected@xdef\@bs@thanks{\@bs@thanks
\protect\footnotetext[\the\c@footnote]{#1}}%
}
\let\@bs@thanks\@empty
\newcommand{\saythanks}{\begingroup
\renewcommand{\thefootnote}{\fnsymbol{footnote}}\@bs@thanks
\endgroup\global\let\@bs@thanks\@empty}
 \makeatother

\iccvfinalcopy

\usepackage[pagebackref=true,breaklinks=true,letterpaper=true,colorlinks,bookmarks=false]{hyperref}

\usepackage[capitalize]{cleveref}
\crefname{section}{Sec.}{Secs.}
\Crefname{section}{Section}{Sections}
\Crefname{table}{Table}{Tables}
\crefname{table}{Tab.}{Tabs.}

\graphicspath{{figures/}}

\DeclareMathOperator{\E}{\mathbb{E}}
\newcommand{\x}{\bm{x}}
\newcommand{\e}{\bm{e}}
\newcommand{\ux}{\bm{u}}

\newcommand{\z}{\bm{z}}

\newcommand{\y}{\bm{y}}

\newcommand{\real}{\mathbb{R}}
\newcommand{\loss}{\mathcal{L}}
\newcommand{\Dloss}{\loss_{D}}
\newcommand{\Gloss}{\loss_{G}}
\newcommand{\lblloss}{\loss_{\textrm{adv}}^{\textrm{lbl}}}
\newcommand{\unlblloss}{\loss_{\textrm{adv}}^{\textrm{unlbl}}}

\newcommand{\fakeloss}{\loss_{\textrm{adv}}^{\textrm{fake}}}
\newcommand{\clsloss}{\loss_{\textrm{cls}}}
\newcommand{\lgce}{l_{\mathrm GCE}}
\newcommand{\best}[1]{$\textbf{#1}$}
\newcommand{\sbest}[1]{$\underline{#1}$}
\newcommand{\ours}{\textbf{Ours}\xspace}
\newcommand{\baseline}{DiffAug CR-GAN\xspace}

\newcommand{\pxy}{(\x, \y)\sim p(\x, \y)}
\newcommand{\pu}{\ux\sim p(\ux)}
\newcommand{\qzy}{(\z, \y)\sim q(\z, \y)}

\def\iccvPaperID{11540} %

\ificcvfinal\fi

\begin{document}

\title{Soft Curriculum for Learning Conditional GANs with Noisy-Labeled and Uncurated Unlabeled Data}
\author[1]{Kai Katsumata\thanks{\tt\small katsumata@nlab.ci.i.u-tokyo.ac.jp}}
\author[1]{Duc Minh Vo}
\author[1,2]{Tatsuya Harada}
\author[1]{Hideki Nakayama}
\affil[1]{The University of Tokyo, Japan}
\affil[2]{RIKEN Center for Advanced Intelligence Project, Japan}

\twocolumn[{%
\renewcommand\twocolumn[1][]{#1}%
\maketitle
\begin{center}
     \centering
       \begin{spacing}{0.5}
     \begin{tabular}{ccccc:cc}
       & \multicolumn{6}{c}{\includegraphics[width=0.7\linewidth]{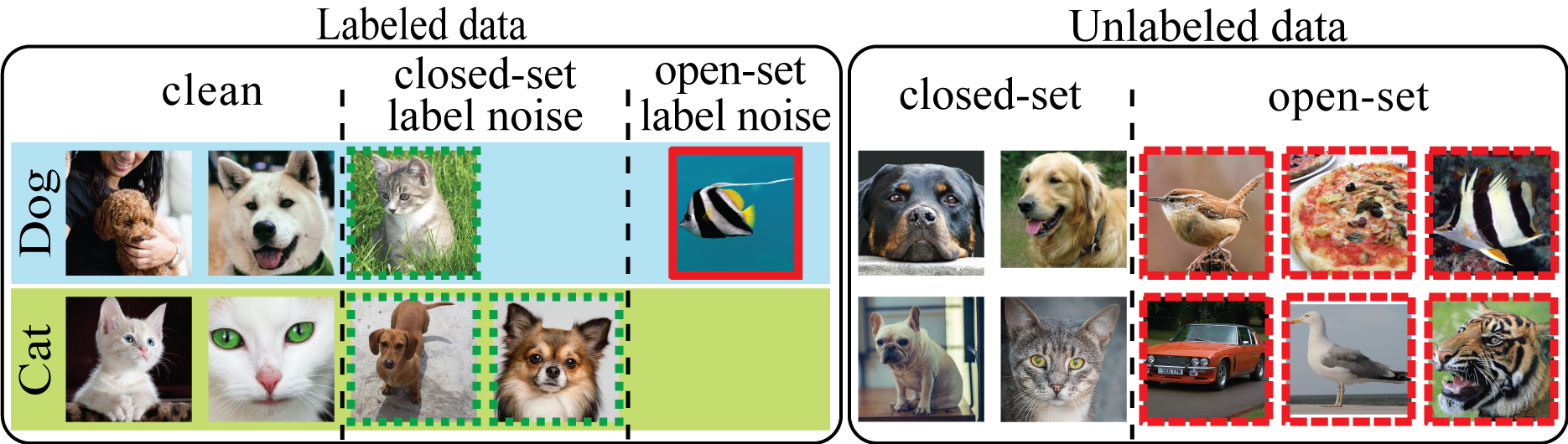} } \\
       {\small (a) Supervised image generation} &&$\checkmark$ &&&& \\ \midrule
       {\!\!\!\small (b) Semi-supervised image generation\!\!\!\!} && $\checkmark$ & & & $\checkmark$ & ($\checkmark$) \\ \midrule
       {\small (c) Noise robust image generation}&& $\checkmark$ & $\checkmark$&&& \\ \midrule
       {\small (d) Ours} &\hspace{0.6em} & \makebox[5em]{$\checkmark$} & \makebox[5em]{$\checkmark$} & \makebox[4em]{$\checkmark$} & \makebox[5em]{$\checkmark$} & \makebox[7em]{$\checkmark$} \\ \midrule
     \end{tabular}
       \end{spacing}
       \vspace{0.1ex}
    \captionsetup{type=figure}
    \caption{We investigate a conditional image generation in which we relax the assumption on training data. 
A dataset consists of labeled and unlabeled data. Labeled data
contains clean samples, closed-set label noise samples whose actual
categories are known classes (green dotted rectangle), and open-set
label noise samples whose actual categories are outside the known
classes (solid red rectangle). Unlabeled data contains closed-set
samples as well as open-set samples whose categories are outside the
known classes (red dashed rectangle). In contrast to previous
assumptions (a,b,c), which allow part of them, our data assumption (d)
generalizes these approaches by integrating a variety of data.  The
$\checkmark$ indicates the full usage, while ($\checkmark$) is partial
usage.  } \label{fig:data_assumption}
\end{center}%
}]\saythanks

\begin{abstract}
  Label-noise or curated unlabeled data is used to compensate for the assumption of clean labeled data in training the conditional generative adversarial network; however, satisfying such an extended assumption is occasionally laborious or impractical. As a step towards generative modeling accessible to everyone, we introduce a novel
  conditional image generation framework that accepts noisy-labeled and
  uncurated unlabeled data during training: (i) closed-set and open-set label noise in
  labeled data and (ii) closed-set and open-set unlabeled data.  To combat it, we
  propose soft curriculum learning, which assigns instance-wise weights for adversarial training while assigning new labels for unlabeled data and correcting wrong labels for labeled data. 
  Unlike popular curriculum learning, which uses a
  threshold to pick the training samples, our soft curriculum controls
  the effect of each training instance by using the weights predicted
  by the auxiliary classifier, resulting in the preservation of useful
  samples while ignoring harmful ones.
Our experiments show that our approach outperforms existing semi-supervised and label-noise robust methods in terms of both quantitative and qualitative performance.
In particular, the proposed approach is able to match the performance of (semi-) supervised GANs even with less than half the labeled data.
\end{abstract}

\section{Introduction}
\label{sec:intro}

Significant breakthroughs~\cite{Mirza2014,Miyato2018b,Zhang2019,Brock2018,ni2022manifold} in class-conditional image generation (cGANs) yield images with high fidelity and diversity; yet they are all trained in a supervised fashion where the training data consists of
carefully labeled samples.
However, the training data for supervised learning requires immense labor-cost, making it difficult to achieve a sophisticated performance.
To deflate the labor-cost in collecting data, semi-supervised~\cite{Lucic2019,katsumata2022ossgan} and label-noise robust~\cite{Kaneko2019,thekumparampil2018robustness} approaches have been investigated.
Despite substantial efforts of semi-supervised cGANs~\cite{Lucic2019,katsumata2022ossgan} to reduce the amount of labeled data, a dataset with a high annotation cost is still required.

In this work, to significantly reduce the data collection and annotation cost, we present a new framework for training cGANs (see \cref{fig:data_assumption}), which utilizes unreliable
labeled data and uncurated unlabeled data. Namely, in this study, we aim to unify the
research directions for training conditional image generation on
imperfect data: annotation quality~\cite{Kaneko2019,thekumparampil2018robustness} and unannotated data~\cite{Lucic2019,katsumata2022ossgan}. In our
realistic data assumption, the dataset consists of two parts: noisy labeled data
(\ie, labeled data with closed-set and open-set label noise) and uncurated unlabeled data
(\ie, unlabeled data with closed-set and open-set samples). Here, closed-set
and open-set label noise mean that the actual labels of samples with
label noise are inside and outside the known category (label) set,
respectively. Closed-set and open-set unlabeled samples also mean
that the actual unknown labels are inside and outside the known category set,
respectively. 
The objective of the new framework is to generate the images with the known categories.
This setting generalizes (i) semi-supervised image generation~\cite{Lucic2019,katsumata2022ossgan} where the labels are reliable, and (ii) label-noise image generation~\cite{Kaneko2019,thekumparampil2018robustness} where labeled data contains only closed-set label noise, and unlabeled data are not available.
Hence, this new data assumption enables the use of personal collection or user-annotated data in conditional image synthesis.

To address the complex data, we propose soft curriculum learning,
which makes clean and fully labeled data from noisy and partially labeled data
while assigning weights to samples for adversarial training.
It eliminates
the harmful samples (\eg, samples failed to assign labels and samples far away from the training categories)
while preserving the useful ones (\eg, samples with proper labels). Motivated by the aim,
we jointly train cGAN and an auxiliary classifier that assigns
clean or new labels to labeled or unlabeled samples, respectively, and
confidences to all real samples. 
Our implicit sample selection mechanism addresses the shortcomings of curriculum learning techniques~\cite{Yu2020,zhang2021flexmatch,gong7465792,cascante2021curriculum}, which potentially retain harmful samples and miss helpful ones because it explicitly uses a predetermined or adaptive threshold.
Consequently, our approach allows curriculum learning to handle noisy labeled and uncurated unlabeled data naturally, resulting in maintaining the number of training samples while reducing the effects of adverse samples.
Since our method is free of the hard selection procedure, we term it as \textit{soft curriculum learning}.

Our comprehensive experiments demonstrate that soft curriculum
learning works well in challenging imperfect datasets containing label noise and
unlabeled data. More precisely, we observe performance gains of our
 method over baselines in terms of Fr\'{e}chet Inception Distance
(FID)~\cite{Heusel2017}, Inception Score
(IS)~\cite{Salimans2016}, $F_{1/8}$, $F_8$~\cite{Sajjadi2018}, and intra-FID (iFID).
Qualitative results also indicate the effectiveness of our
method in terms of image fidelity and diversity.

In summary, our main contributions are as follows:
\begin{enumerate}
\item We introduce a new problem: conditional image generation trained on datasets
that consists of labeled data with closed-set and open-set label noise and
unlabeled data composed of closed-set and open-set samples.
\item We develop a soft curriculum technique for correcting wrong labels and assigning
temporal labels while weighting importances of each instance by
employing an additional classifier trained jointly.
\item We consistently demonstrate the effectiveness of our method in experiments on a variety of GAN architectures (\ie, projection- and classifier-based cGANs) and datasets.
  Note that recent attempts at limited data employ only a projection GAN.
\end{enumerate}

\section{Related work}
\noindent \textbf{Conditional image generation with imperfect data.}
One of the prominent research directions in image generation is to
build a training framework without requiring large and curated
datasets.
Semi-supervised learning approaches~\cite{Ting2019,Lucic2019,katsumata2022ossgan} explore cGANs in partially labeled data.
Introducing an additional classifier enables a discriminator to train on labeled real data. OSSGAN~\cite{katsumata2022ossgan} considers
a more practical scenario 
where the labeled and unlabeled data do not share the label space, and it proposes entropy regularization to identify open-set samples smoothly.
Robust learning for image generation~\cite{Kaneko2019,thekumparampil2018robustness} aims to learn a clean conditional distribution even
when labels are noisy by modeling a noise transition. 
In this study, we extend these directions to a real-world scenario. Our setting relaxes the assumption of label reliability in a semi-supervised fashion and
allows robust learning to exploit open-set label noise and unlabeled data.

\noindent \textbf{Semi-supervised and robust learning in image
  recognition.}  Image recognition also remains the issue that supervised
learning requires datasets, which are difficult and sometimes
impossible to collect, \ie cleanly labeled large-scale datasets. To address the issue, two
popular frameworks (\ie, semi-supervised~\cite{rasmus2015semi,hataya2019unifying} and label-noise
robust learning~\cite{natarajan2013learning}) have been explored in recent decades.
Recent attempts address a more realistic scenario where the categories of samples are not bounded by
the known categories. Open-set semi-supervised learning~\cite{saito2021openmatch,Yu2020,Luo2021} involves unlabeled data containing samples with categories unseen in labeled data, aiming to classify closed-set samples precisely while rejecting open-set samples.
Learning methods robust to closed-set and open-set label noise~\cite{albert2022addressing,sachdeva2021evidentialmix,yao2021jo,wang2018iterative}  generalize
methods that only consider closed-set noise~\cite{angluin1988learning,natarajan2013learning}. %
In this study, we attempt to unify these research directions that are independently addressed in conditional image synthesis.

\section{Problem statement}

We present a novel training setting for data-efficient conditional image generation that leverages noisy labeled data and
uncurated unlabeled data.
For $K$-class conditional image generation, let $\mathcal{D}_l = \{(\x_i, \y_i)\}^{n_l}_{i=1}$ be the noisy labeled training set consisting of $n_l$ labeled samples, where
a $d$-dimensional instance $\x_i \in \real^{d}$ and its corresponding noisy label $\y_i \in \mathcal{Y}$ that are sampled from labeled data distribution $p(\x, \y)$ : $(\x_i, \y_i)\sim p(\x, \y)$. The noisy label space $\mathcal{Y} = \{\e^{(1)},\ldots,\e^{(K-1)},\e^{(K)}\}$ consists of the standard basis vectors of the $K$-dimensional space.
The clean label space $\bar{\mathcal{Y}} = \mathcal{Y}\cup\{\textrm{open-set classes}\}$ is inaccessible.
Let $\mathcal{D}_u = \{\ux_i\}^{n_u}_{i=1}$ be an uncurated unlabeled training set having ${n_u}$ samples, where an instance $\ux_i\in \real^{d}$ is sampled from unlabeled data distribution
$p(\ux)$ : $\ux_i\sim p(\ux)$. Unlabeled data also includes both closed-set and open-set samples.
The goal of the conditional image generation is to model the true distribution without label noise via 
a generator $G$ and a discriminator $D$.
The generator $G$ generates samples $G(\z,\y)$ from a latent vector $\z \in \real^{d_z}$ and a conditioning label $\y$
drawn from a prior distribution $\qzy = q(\z)q(\y)$, where $q(\z)$ is typically the standard Gaussian distribution
and $q(\y)$ is the uniform distribution over $\mathcal{Y}$. 
The discriminator $D$ aims to identify fake samples $(G(\z, \y), \y)$ from real samples $(\x, \y)$.

Before formulating our method, we introduce a supervised cGAN model.
The conditional GANs for a fully and cleanly labeled dataset optimize the losses $\Dloss$ and $\Gloss$ for the discriminator and the generator, respectively:
\begin{align}
  \Dloss = & \E_{\pxy} [  f_{D}(-D(\x, \y))] \nonumber\\
  & + \E_{\qzy} [ f_{D}(D(G(\z, \y), \y))],\label{eq:supervised_dloss}\\
\Gloss = & \E_{\qzy} [-D(G(\z, \y), \y)],\label{eq:supervised_gloss}
\end{align}
where a hinge loss~\cite{Lim2017,Tran2017} for the discriminator $f_D(\cdot) = \max(0, 1+\cdot)$.
Updates of the generator and discriminator parameters with $\Gloss$ and $\Dloss$ alternately make a generator that generates indistinguishable samples and a discriminator that distinguishes fake and real samples well.
To present our method, we customize the cGANs (\cref{eq:supervised_dloss,eq:supervised_gloss}).

Although SoTA cGANs achieve outstanding performance,  
the absence of a dataset with sufficient quantity and reliable labels leads to poor performance and training instability.
Difficulties in training on a dataset with limited quantity and quality are how to improve the stability of the training and how to estimate appropriate labels to unlabeled data under noisy labels.
To overcome the difficulties, we consider a technique that assigns labels while handling label noise based on curriculum learning and robust learning. %

\section{Method}\label{sec:method}

\begin{figure}[tb]
\centering
\includegraphics[width=1\linewidth]{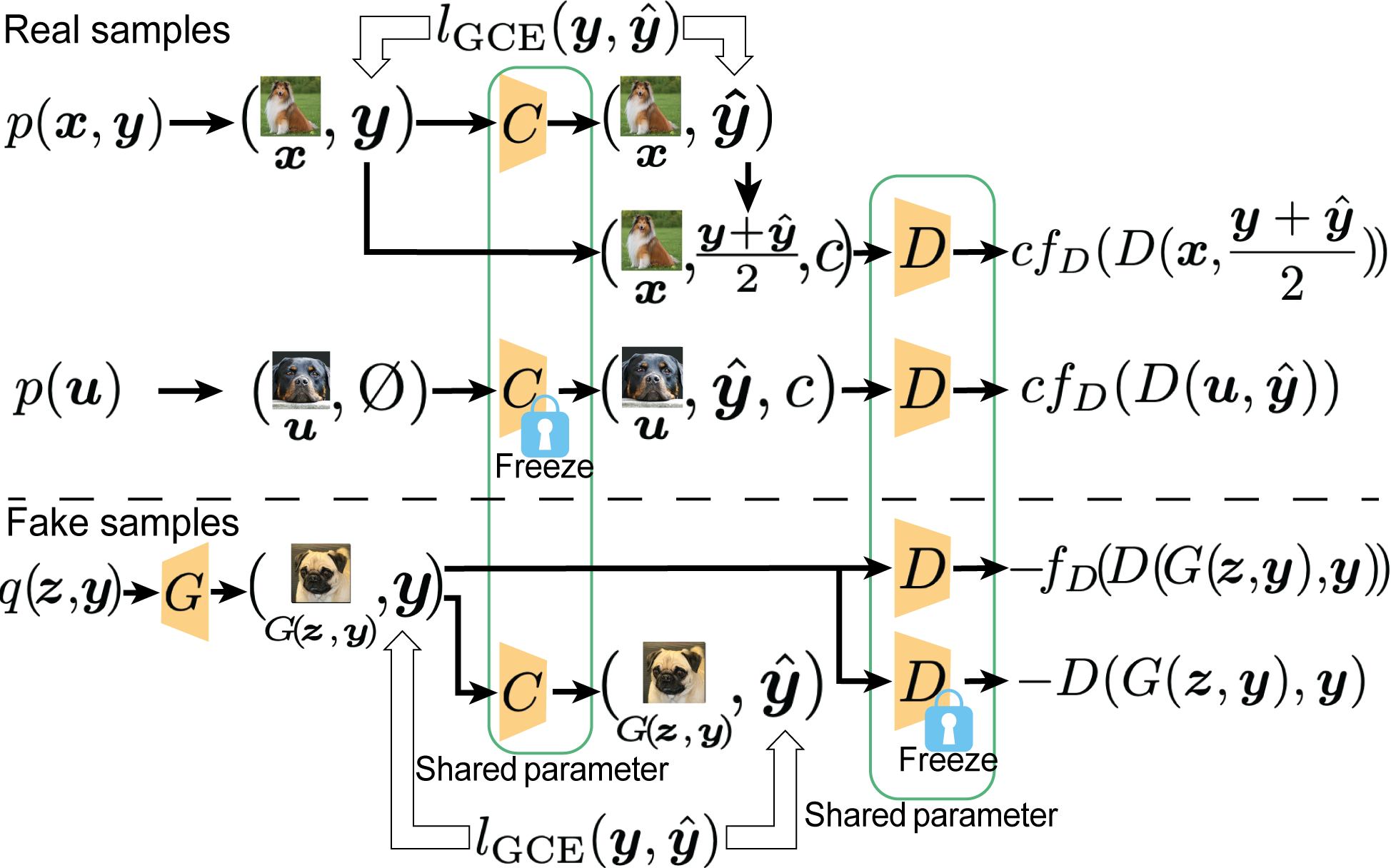}
\caption{Overview of the proposed method. The auxiliary classifier is trained with the classification loss $\lgce$ (\cref{eq:gceloss}). It corrects wrong labels in labeled samples by $C(\x)$, 
  assigns labels to unlabeled samples by $C(\ux)$, and distributes confidences $c$
  for the discriminator optimization (\cref{eq:confidence}). The discriminator is trained with the adversarial loss for labeled
data, unlabeled data, and fake data (\cref{eq:lblloss,eq:unlblloss,eq:fakeloss}). Zoom in for best view.}\label{fig:method}
\end{figure}

\noindent \textbf{Intuitive idea}.
Curriculum learning~\cite{Yu2020,zhang2021flexmatch,gong7465792,cascante2021curriculum} filters out adverse samples from the dataset, aiming to train a model on only useful samples. However, since curriculum learning employs explicit thresholds, it does not leverage the feature of ignored samples, resulting in shrinking training datasets.
Furthermore, curriculum learning methods~\cite{Yu2020} for semi-supervised learning maintain label noise.

To overcome these flaws, we consider a safer way for learning cGANs on noisy data, aiming to reduce the adverse effect of misclassification while maintaining the amount of training data.
Therefore, we have to achieve three objects:
handling label noise containing open-set noise; handling unlabeled data including open-set samples; and eliminating samples causing negative effects from both labeled and unlabeled data. 
Our main idea is to make clean data from noisy labeled and uncurated unlabeled data and to control the effects of each instance tolerantly.
Our method can train the discriminator on all samples via the instance-wise weight distribution, label correction, and label assignment (\cref{fig:method}), 
unlike curriculum learning, which picks unlabeled samples and trains a model on all the labeled data and the selected unlabeled data. 
Our instance-wise weighting mechanism leads to reducing the negative effects of label noise in labeled data by assigning small weights for samples that could not be corrected by the auxiliary classifier or are open-set.

\noindent \textbf{Overall concept}.
In addition to a generator $G: \real^{d_z} \times \mathcal{Y} \to \real^{d} $ and a discriminator $D: \real^{d} \times \Delta^{K-1} \to \real$, we employ a classifier $C: \real^{d} \to \Delta^{K-1}$ where $\Delta^{K-1}$ is a probability simplex whose vertices are in $\mathcal{Y}$.
To extend the above loss function (\cref{eq:supervised_dloss,eq:supervised_gloss}) into our setting, we introduce discriminator losses for noisy
labeled data and uncurated unlabeled data $\lblloss, \unlblloss$ and an auxiliary classifier loss $\clsloss$.
Our approach can be divided into four key components: training a robust auxiliary classifier, assigning new labels to unlabeled data,
correcting labels for labeled data, and weighting loss for real data (\ie, both labeled and unlabeled data).
For involving noisy labeled and unlabeled data, we optimize the loss functions $\Dloss$ and $\Gloss$:
\begin{align}
\Dloss = & \lblloss + \unlblloss + \fakeloss + \lambda \clsloss\label{eq:dloss}, \\
\Gloss = & \E_{\qzy} [-D(G(\z, \y), \y)],\label{eq:gloss}
\end{align}
where $\lambda$ is a balancing parameter between the adversarial loss and the classification loss.
We use the discriminator loss for fake data in the same as the supervised way:
\begin{align}
  \fakeloss = \E_{\qzy} [ f_{D}(D(G(\z, \y), \y))].\label{eq:fakeloss}
\end{align}

Soft curriculum is an instance-wise weighting framework for discriminator training, which aims to assign small weights to harmful or irrelevant samples (\eg, wrongly labeled closed-set samples and open-set samples)
and large weights to helpful samples (\eg, correctly labeled samples).

\noindent \textbf{Robust training of auxiliary classifier}.
We employ an auxiliary classifier for label assignment and correction (the details in a later paragraph). 
In training the classifier, besides real labeled data, we also use generated samples to increase the training samples.
Incorporating generated samples into the training may prevent memorizing training samples (\ie, overfitting). 
The classification loss is given by:
\begin{align}
  \clsloss = & \E_{\pxy} [\lgce(C(\x), \y)]\nonumber \\
                  & + \E_{\qzy} [\lgce(C(G(\z, \y)), \y)].\label{eq:clsloss}
\end{align}
For robust classification with label noise,
we use the generalized cross entropy~\cite{zhang2018generalized}, which is the generalization of the mean absolute error (MAE)~\cite{ghosh2017robust} and the cross entropy. %
The loss of the generalized cross entropy is given by
\begin{align}
  \lgce(\x, \y) = \frac{1 - (\x^{\mathsf{T}} \y)^{q}}{q},\label{eq:gceloss}
\end{align}
where, the hyperparameter $q \in [0, 1]$ controls the trade-off between optimization and noise robustness.
When $q = 1$, it is equivalent to the MAE, which is robust to label noise but difficult to optimize.
When $q = 0$, it is equivalent to the cross entropy loss, which can be optimized easily. %
The discriminator and classifier share the feature extractor
to extract features efficiently.
We use the classifier prediction for label assignment for unlabeled data and label correction for labeled data.

\noindent \textbf{Label assignment for unlabeled data}.
To assign new labels to unlabeled data, we take classifier's softmax outputs $\hat{\y} = C(\ux)$ as a condition in discriminator inputs.
We use soft labels (\ie probability vector) for the robustness to classification errors and open-set samples instead of hard labels. 
Soft labels prevents the discriminator inputs from wrong labels with the classifier mistake
because soft labels assign a small probability to the correct class and avoid assigning a probability of 1 to the wrong class.

\noindent \textbf{Label correction for labeled data}.
To correct noisy labels for labeled data, we take the interpolation between a given label and a predicted label: $(\y + \hat{\y})/2$, before feeding labels into the discriminator where $\hat{\y} = C(\x)$.
Since some samples have proper labels depending on the label noise ratio, overwriting the given labels loses helpful information about samples with correct labels.
We use the simple average because the average weighted with confidence may amplify the negative effects of wrong predictions.
While we use predicted labels for inputs of the discriminator to real labeled and unlabeled samples, we maintain labels for generated samples because their labels are already proper.

\noindent \textbf{Confidence assignment}. To focus on helpful samples, we quantify the sample-wise importance in the discriminator training via classifier predictions.
The discriminator losses for labeled and unlabeled data are defined by
\begin{align}
  \lblloss &=  \E_{\pxy} [ c f_{D}(-D(\x, (\y + \hat{\y})/2))], \label{eq:lblloss}\\
  \unlblloss &= \E_{\pu} [c f_{D}(-D(\ux, \hat{\y}))],\label{eq:unlblloss}
\end{align}  
where $\hat{\y} = C(\x)$ and $\hat{\y} = C(\ux)$ are the softmax output of the classifier, and the confidence in the soft curriculum $c \in [0, 1]$ is the normalized entropy of the classifier prediction:
\begin{align}
  c = 1 - \frac{\sum_{\hat{y}_i \in \hat{\y}} \hat{y}_i \log \hat{y}_i}{\log K}.\label{eq:confidence}
\end{align}
Here, it assigns large $c$ for samples with high confidence and small $c$ for samples with low confidence.

\begin{table*}[tb]
\centering
\caption{Average and standard deviation of $F_{8}$, $F_{1/8}$, FID, Inception score (IS), and iFID over three trials on TinyImageNet with 150 closed-set
  classes, 20\% labeled samples, and 10\% label noise. We compare our proposed method with 15 baselines.
Our method yields better performance (\ie, the higher $F_{8}$, $F_{1/8}$, and IS and lower FID and iFID) and consistent performance
  (small standard deviation). The best results are highlighted in \best{bold}, and
  the second best results are \sbest{underlined}.
  }\label{tb:tiny150_10_20}
\resizebox{0.95\linewidth}{!}{
\begin{tabular}{lccccc}\toprule
& $F_{8}\uparrow$ & $F_{1/8}\uparrow$ & FID$\downarrow$ & IS$\uparrow$ & iFID$\downarrow$\\\midrule
\baseline~\cite{Zhao2020} & 0.9341 $\pm$ .0103 & 0.9669 $\pm$ .0034 & 41.6848 $\pm$ 1.0075 & 12.0270 $\pm$ 0.3451  & 227.2077 $\pm$ 3.3538\\
RandomGAN & 0.6908 $\pm$ .0310 & 0.8061 $\pm$ .0492 & 84.2262 $\pm$ 9.7936 & \phantom{0}7.6780 $\pm$ 0.6785 & 312.8149 $\pm$ 6.1245\\
SingleGAN & 0.9374 $\pm$ .0009 & \sbest{0.9761} $\pm$ .0018 & 35.5989 $\pm$ 1.5018 & 12.3043 $\pm$ 0.2951 & 233.8048 $\pm$ 4.3930 \\
$S^3$GAN~\cite{Lucic2019} & 0.9287 $\pm$ .0027 & 0.9667 $\pm$ .0031 & 39.8652 $\pm$ 1.2017 & 12.1443 $\pm$ 0.2344  & 223.5165 $\pm$  0.5562  \\
OSSGAN~\cite{katsumata2022ossgan} & 0.8954 $\pm$ .0119 & 0.9598 $\pm$ .0029 & 46.9769 $\pm$ 3.0722 & 10.8745 $\pm$ 0.4495 & 236.6557 $\pm$  5.0004\\
CurriculumGAN & 0.9146 $\pm$ .0128 &  0.9388 $\pm$ .0144 &  34.4142 $\pm$ 0.6545 & \sbest{13.3153} $\pm$ 0.6545 & \sbest{217.9899} $\pm$ 1.5723\vspace{-0.6em} \\
\multicolumn{6}{c}{\dotfill} \\
reRandomGAN & 0.4890 $\pm$ .0396 & 0.7653 $\pm$ .0154 & 88.9622 $\pm$ 3.7217 &  \phantom{0}6.8242 $\pm$ 0.5130 & 317.4159 $\pm$ 2.2235 \\
reSingleGAN & 0.8969 $\pm$ .0047 & 0.9422 $\pm$ .0099 & 36.2851 $\pm$ 1.3121 & 12.4421 $\pm$ 0.3234 & 237.1689 $\pm$ 2.2875 \\
re$S^3$GAN & 0.9089 $\pm$ .0070 & 0.9476 $\pm$ .0024 & 37.4676 $\pm$ 0.7783 & 13.0772 $\pm$ 0.2206 & 221.3113 $\pm$ 0.6992 \\
reOSSGAN & 0.8745 $\pm$ .0037 & 0.9320 $\pm$ .0044 & 40.1548 $\pm$ 1.1753 & 12.1081 $\pm$ 0.1531 & 229.1839 $\pm$ 1.6075 \vspace{-0.6em}\\
\multicolumn{6}{c}{\dotfill} \\
rcDiffAugCRGAN & 0.9332 $\pm$ .0044 & 0.9617 $\pm$ .0078 & 43.5950 $\pm$ 2.2703 & 11.8126 $\pm$ 0.4097 & 226.1654 $\pm$  4.4462\\
rcRandomGAN & 0.7466 $\pm$ .0298 & 0.8801 $\pm$ .0312 & 69.7574 $\pm$ 4.8421 & \phantom{0}7.5598 $\pm$ 0.9622 & 293.7392 $\pm$  5.9841\\
rcSingleGAN & \sbest{0.9409} $\pm$ .0072 & 0.9743 $\pm$ .0026 & \sbest{34.1262} $\pm$ 1.3978 & 12.9476 $\pm$ 0.3931 & 223.1789 $\pm$  3.8244\\
rc$S^3$GAN & 0.9258 $\pm$ .0072 & 0.9661 $\pm$ .0056 & 42.0012 $\pm$ 2.1783 & 12.0116 $\pm$ 0.3488  & 228.4053 $\pm$  4.8632\\
rcOSSGAN & 0.9281 $\pm$ .0082 & 0.9692 $\pm$ .0006 & 42.0705 $\pm$ 1.1632 & 12.0458 $\pm$ 0.2670 & 227.5382 $\pm$ 2.3760\vspace{-0.6em}\\
\multicolumn{6}{c}{\dotfill} \\ 
\ours & \best{0.9581} $\pm$ .0063 & \best{0.9789} $\pm$ .0003 & \best{29.6607} $\pm$ 0.4979 & \best{14.7235} $\pm$ 0.3509 & \best{206.6937} $\pm$ 2.1925 \\\bottomrule
\end{tabular}}
\end{table*}

\noindent \textbf{Implementation details.}
In the experiments on the Tiny ImageNet~\cite{le2015tiny} datasets at $64\times 64$ resolution, we use a minibatch size of $1024$, the latent dimension of $100$, and
the learning rates of $1\!\times\!10^{-4}\!$ and $4\times\!10^{-4}\!$ for the generator and the discriminator, respectively.
In the experiments on the ImageNet~\cite{Olga2015} and WebVision~\cite{li2017webvision} datasets at $128\times 128$ resolution, we have a minibatch size of $256$, the latent dimension of $120$, and
learning rates of $5\!\times\!10^{-5}\!$ and $2\!\times\!10^{-4}\!$ for the generator and the discriminator, respectively.
We update a discriminator in two steps per iteration. We train the auxiliary classifier with the same learning rate as the discriminator.
We select a parameter $\lambda$ in the preliminary experiments with the 150-class TinyImageNet dataset and set $0.1$ for all the experiments.
The parameter $q$ in generalized cross entropy is $0.7$, which is the default value in \cite{zhang2018generalized}.

\section{Experiments}

\noindent \textbf{Datasets.}
For the comprehensive evaluation, we perform experiments on TinyImageNet~\cite{TinyImageNet}, ImageNet~\cite{Olga2015}, and WebVision~\cite{li2017webvision} datasets.
We construct partially labeled datasets consisting of noisy labeled and uncurated unlabeled samples to benchmark our method.
We use four variables that control a dataset configuration: the ratio of label noise, the number of closed-set classes, the labeled sample ratio,
and the usage ratio. For the WebVision dataset, we omit the procedure for injecting label noise since it already contains label noise.
To raise the open-set label noise, we first shuffle the labels by the ratio of label noise. 
We change a label to another label uniformly with the probability of the ratio of label noise. 
The label transition is run among all the classes.
Second, we divide the fully labeled dataset with flipped labels into a part of closed-set classes and a part of open-set classes.
The rest of the classes subtracted the number of closed-set classes from 1000 classes are considered as open-set classes.
Since label noise is brought before separation into closed-set and open-set classes, the subset for closed-set classes contains both open-set and closed-set label noise.
Then, we take a subset of closed-set samples according to the labeled sample ratio as labeled data, and we take the remaining closed-set samples as unlabeled data.
Finally, we extract unlabeled samples from open-set class samples with the usage ratio and concatenate them with unlabeled samples that come from closed-set samples. 
We use the usage ratio of 100\%, if not otherwise specified.

\noindent \textbf{Compared methods.}
We use CR-BigGAN~\cite{zhang2019consistency} with DiffAugment~\cite{Zhao2020} (\baseline)
as a base architecture, and we build all the compared methods on it.
We compare the proposed method (\ours), with \baseline~\cite{Brock2018}, RandomGAN, SingleGAN, $S^3$GAN~\cite{Lucic2019}, OSSGAN~\cite{katsumata2022ossgan}, and CurriculumGAN.
RandomGAN is a naive baseline and assigns labels to unlabeled samples by picking a label from $\y \in \mathcal{Y}$ with equal probability.
SingleGAN is another simple baseline and assigns constant labels $[1/K,\ldots,1/K]^{\textsf{T}}$ to all unlabeled samples without considering their content.
CurriculumGAN uses a curriculum learning for semi-supervised learning by following \cite{Yu2020} instead of our soft curriculum.
For further comparison, we introduce two types of extended baselines (\ie, relabeling and rcGAN~\cite{Kaneko2019}). The extended relabeling baselines are denoted by the prefix `re' correct labels of labeled samples by using \cref{eq:lblloss} and predicted labels $\hat{\y} = C(\x)$ for labeled samples. 
The methods with the prefix `rc' have rcGAN, which is a technique for robust learning with label noise.
The details of the compared methods are given in the supplementary material.

\begin{table*}[tb]
\centering
\caption{Ablation study on Tiny ImageNet with 150 closed-set classes, 20\% labeled samples, and 30\%/50\% labeled noise.
  AB1 is the method without generalized cross entropy. AB2 is the method without curriculum learning. AB3 is the
  method without curriculum for labeled data.
}\label{tb:ablation}
\resizebox{0.85\linewidth}{!}{\begin{tabular}{lcccccccccc}\toprule
& \multicolumn{5}{c}{30\% label noise} & \multicolumn{5}{c}{50\% label noise} \\ \cmidrule(lr){2-6} \cmidrule(lr){7-11}
& $F_{8}\uparrow$ & $F_{1/8}\uparrow$ & FID$\downarrow$ & IS$\uparrow$ & iFID$\downarrow$ & $F_{8}\uparrow$ & $F_{1/8}\uparrow$ & FID$\downarrow$ & IS$\uparrow$ & iFID$\downarrow$  \\\midrule
AB1 & 0.8874 & 0.9615 & 36.2120 & 12.3104 & 232.8597 & 0.8910 & 0.9427 & \sbest{35.5164} & \sbest{12.2659} & 245.5752\\
AB2 & 0.9092 & 0.9619 & 39.7125 & 11.6496 & 236.4422 & \sbest{0.9131} & \sbest{0.9671} & 40.9006 & 10.7329 & 253.2198 \\
AB3 & \sbest{0.9145} & \sbest{0.9625} & \sbest{31.2353} & \sbest{13.5164} & \sbest{222.1403} & 0.8322 & 0.9517 & 35.6693 & 11.4738 & \sbest{241.6102}\\
\ours & \best{0.9238} & \best{0.9664} & \best{30.5527} & \best{14.0052} & \best{221.6443} & \best{0.9492} &  \best{0.9743} & \best{33.0788} & \best{12.3833} & \best{238.7180}\\\bottomrule
\end{tabular}}
\end{table*}

\begin{table}[tb]
\centering
\caption{Quantitative comparison on ImageNet with closed-set 100 classes, 5\% labeled data, 10\% label-noise.
Our method outperforms the baselines in terms of all metrics.
}\label{tb:imagenet100_10_05_010} 
\resizebox{1\columnwidth}{!}{\begin{tabular}{lccccc}\toprule
& $F_{8}\uparrow$ & $F_{1/8}\uparrow$ & FID$\downarrow$ & IS$\uparrow$ & iFID$\downarrow$\\\midrule
\baseline & 0.8526 & 0.7430&  82.8757&  14.6339 & \sbest{256.1464} \\
RandomGAN & 0.7479 & \sbest{0.8783}&  70.9336&  15.0161 & 300.5579 \\
SingleGAN & 0.6599 & 0.8349&  77.8994&  14.0210 & 310.9954 \\
$S^3$GAN & 0.8429  & 0.8758  &  \sbest{65.9445} &  16.1675 & 264.8367 \\
OSSGAN & \sbest{0.8959} & 0.8453&  68.4343&  \sbest{17.4661} & 284.4511 \\
\ours & \best{0.9443} & \best{0.9430} & \best{57.1299} & \best{22.3548} & \best{219.5597} \\ \bottomrule
\end{tabular}}
\end{table}

\begin{figure*}[tb]
\centering
\includegraphics[width=1\linewidth]{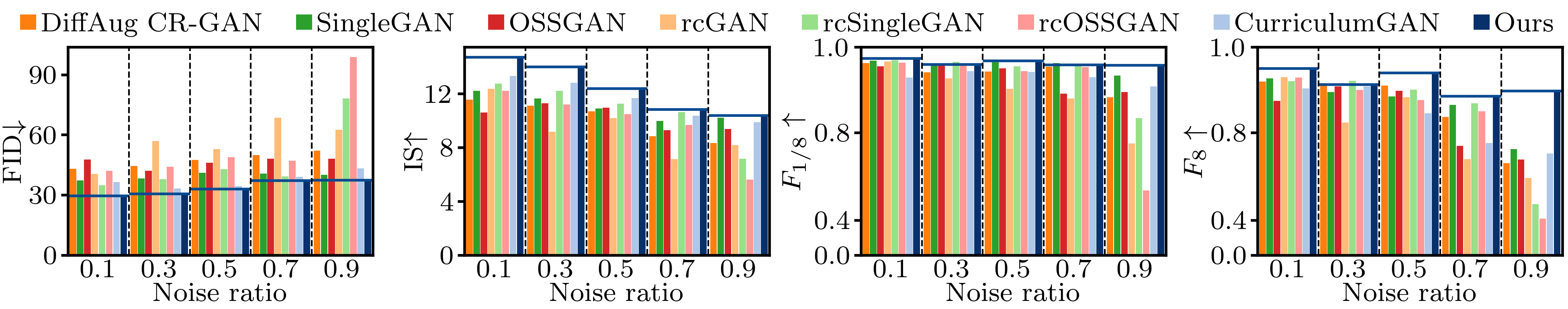}
\caption{Quantitative comparison over different label noise ratios. We report the results of 
the experiments on the TinyImageNet dataset with 150 classes, 20\% labeled data, and label noise ratio of $\{10\%, 30\%, 50\%, 70\%, 90\%\}$.
  We compare the methods over datasets with different label noise ratios. %
The blue lines indicate
  the results of the proposed method. 
Our method considerably outperforms baselines on difficult datasets (\ie, large noise ratio).
}\label{fig:labelnoise}
\end{figure*}

 \begin{figure}[tb]
  \centering
    \bgroup 
    \def\arraystretch{0.2} 
    \setlength\tabcolsep{0.2pt}
    \begin{tabular}{ccccccc}
\includegraphics[width=0.16\linewidth]{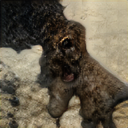} &
\includegraphics[width=0.16\linewidth]{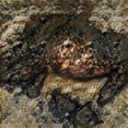} &
\includegraphics[width=0.16\linewidth]{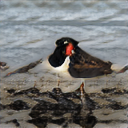} &
\includegraphics[width=0.16\linewidth]{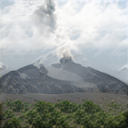} &
\includegraphics[width=0.16\linewidth]{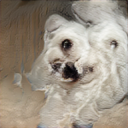} &
\includegraphics[width=0.16\linewidth]{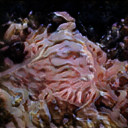} &
\\\multicolumn{6}{c}{\baseline~\cite{Brock2018}} \\ \\
\includegraphics[width=0.16\linewidth]{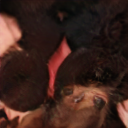} &
\includegraphics[width=0.16\linewidth]{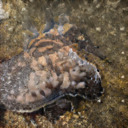} &
\includegraphics[width=0.16\linewidth]{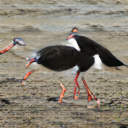} &
\includegraphics[width=0.16\linewidth]{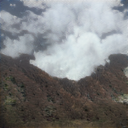} &
\includegraphics[width=0.16\linewidth]{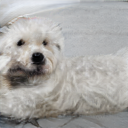} &
\includegraphics[width=0.16\linewidth]{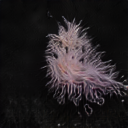}
\\\multicolumn{6}{c}{RandomGAN} \\ \\
\includegraphics[width=0.16\linewidth]{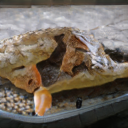} &
\includegraphics[width=0.16\linewidth]{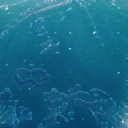} &
\includegraphics[width=0.16\linewidth]{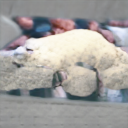} &
\includegraphics[width=0.16\linewidth]{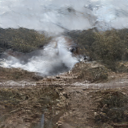} &
\includegraphics[width=0.16\linewidth]{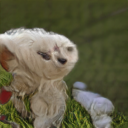} &
\includegraphics[width=0.16\linewidth]{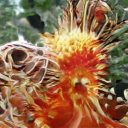}
\\\multicolumn{6}{c}{SingleGAN} \\ \\
\includegraphics[width=0.16\linewidth]{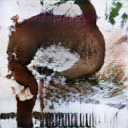} &
\includegraphics[width=0.16\linewidth]{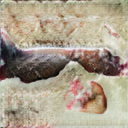} &
\includegraphics[width=0.16\linewidth]{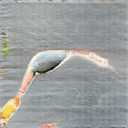} &
\includegraphics[width=0.16\linewidth]{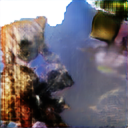} &
\includegraphics[width=0.16\linewidth]{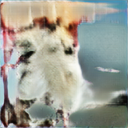} &
\includegraphics[width=0.16\linewidth]{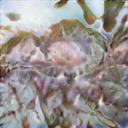}
\\\multicolumn{6}{c}{$S^3$GAN~\cite{Lucic2019}} \\ \\
\includegraphics[width=0.16\linewidth]{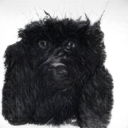} &
\includegraphics[width=0.16\linewidth]{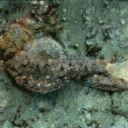} &
\includegraphics[width=0.16\linewidth]{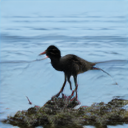} &
\includegraphics[width=0.16\linewidth]{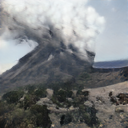} &
\includegraphics[width=0.16\linewidth]{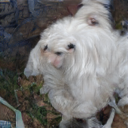} &
\includegraphics[width=0.16\linewidth]{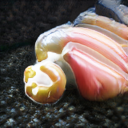} &
\\\multicolumn{6}{c}{OSSGAN~\cite{katsumata2022ossgan}} \\ \\ 
\includegraphics[width=0.16\linewidth]{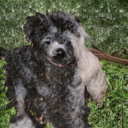} &
\includegraphics[width=0.16\linewidth]{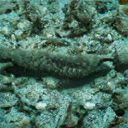} &
\includegraphics[width=0.16\linewidth]{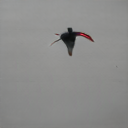} &
\includegraphics[width=0.16\linewidth]{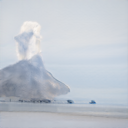} &
\includegraphics[width=0.16\linewidth]{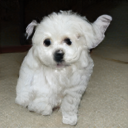} &
\includegraphics[width=0.16\linewidth]{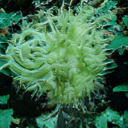}
\\\multicolumn{6}{c}{\ours}
    \end{tabular}\egroup
    \caption{Visual comparison of class-conditional image synthesis results on ImageNet.
      Our method produces plausible images while respecting the given condition.
    }\label{fig:gen_imagenet200_10_05_010_2} 
\end{figure}

 \begin{figure*}[tb]
  \centering
    \bgroup 
    \def\arraystretch{0.2} 
    \setlength\tabcolsep{0.2pt}
    \begin{tabular}{ccccccccccccccccc}
\multicolumn{5}{c}{\baseline} &  &\multicolumn{5}{c}{OSSGAN} &  &\multicolumn{5}{c}{\ours} \\
\includegraphics[width=0.064\linewidth]{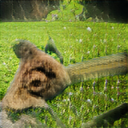} &
\includegraphics[width=0.064\linewidth]{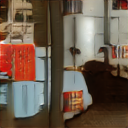} &
\includegraphics[width=0.064\linewidth]{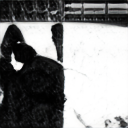} &
\includegraphics[width=0.064\linewidth]{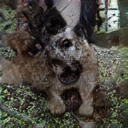} &
\includegraphics[width=0.064\linewidth]{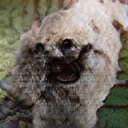} &
\phantom{0}&
\includegraphics[width=0.064\linewidth]{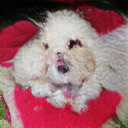} &
\includegraphics[width=0.064\linewidth]{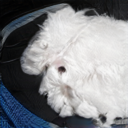} &
\includegraphics[width=0.064\linewidth]{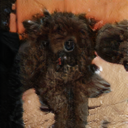} &
\includegraphics[width=0.064\linewidth]{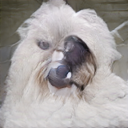} &
\includegraphics[width=0.064\linewidth]{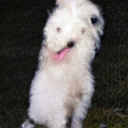} &
\phantom{0}&
\includegraphics[width=0.064\linewidth]{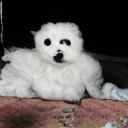} &
\includegraphics[width=0.064\linewidth]{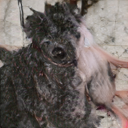} &
\includegraphics[width=0.064\linewidth]{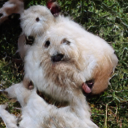} &
\includegraphics[width=0.064\linewidth]{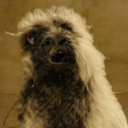} &
\includegraphics[width=0.064\linewidth]{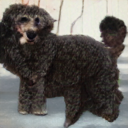} \\
\includegraphics[width=0.064\linewidth]{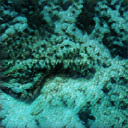} &
\includegraphics[width=0.064\linewidth]{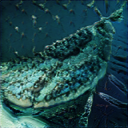} &
\includegraphics[width=0.064\linewidth]{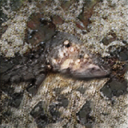} &
\includegraphics[width=0.064\linewidth]{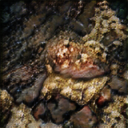} &
\includegraphics[width=0.064\linewidth]{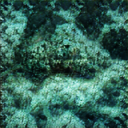} &
&
\includegraphics[width=0.064\linewidth]{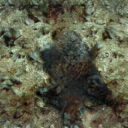} &
\includegraphics[width=0.064\linewidth]{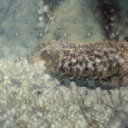} &
\includegraphics[width=0.064\linewidth]{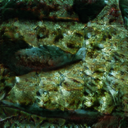} &
\includegraphics[width=0.064\linewidth]{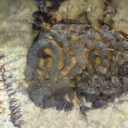} &
\includegraphics[width=0.064\linewidth]{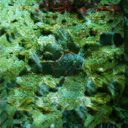} &
&
\includegraphics[width=0.064\linewidth]{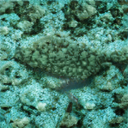} &
\includegraphics[width=0.064\linewidth]{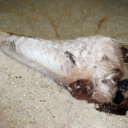} &
\includegraphics[width=0.064\linewidth]{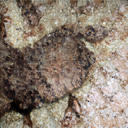} &
\includegraphics[width=0.064\linewidth]{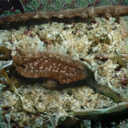} &
\includegraphics[width=0.064\linewidth]{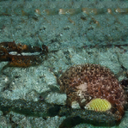} \\
\includegraphics[width=0.064\linewidth]{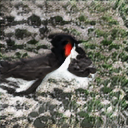} &
\includegraphics[width=0.064\linewidth]{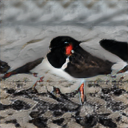} &
\includegraphics[width=0.064\linewidth]{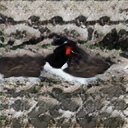} &
\includegraphics[width=0.064\linewidth]{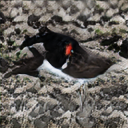} &
\includegraphics[width=0.064\linewidth]{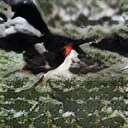} &
&
\includegraphics[width=0.064\linewidth]{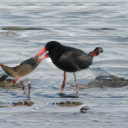} &
\includegraphics[width=0.064\linewidth]{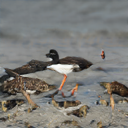} &
\includegraphics[width=0.064\linewidth]{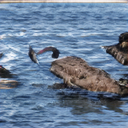} &
\includegraphics[width=0.064\linewidth]{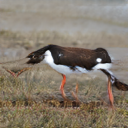} &
\includegraphics[width=0.064\linewidth]{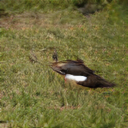} &
&
\includegraphics[width=0.064\linewidth]{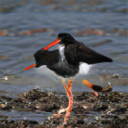} &
\includegraphics[width=0.064\linewidth]{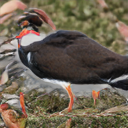} &
\includegraphics[width=0.064\linewidth]{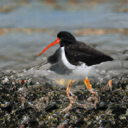} &
\includegraphics[width=0.064\linewidth]{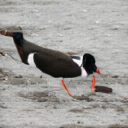} &
\includegraphics[width=0.064\linewidth]{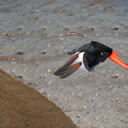} \\
\includegraphics[width=0.064\linewidth]{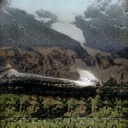} &
\includegraphics[width=0.064\linewidth]{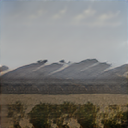} &
\includegraphics[width=0.064\linewidth]{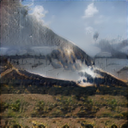} &
\includegraphics[width=0.064\linewidth]{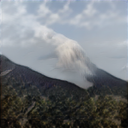} &
\includegraphics[width=0.064\linewidth]{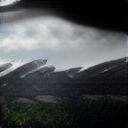} &
&
\includegraphics[width=0.064\linewidth]{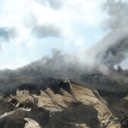} &
\includegraphics[width=0.064\linewidth]{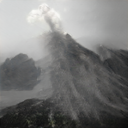} &
\includegraphics[width=0.064\linewidth]{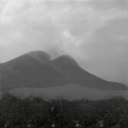} &
\includegraphics[width=0.064\linewidth]{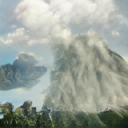} &
\includegraphics[width=0.064\linewidth]{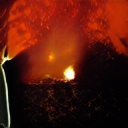} &
&
\includegraphics[width=0.064\linewidth]{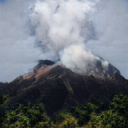} &
\includegraphics[width=0.064\linewidth]{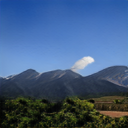} &
\includegraphics[width=0.064\linewidth]{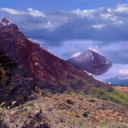} &
\includegraphics[width=0.064\linewidth]{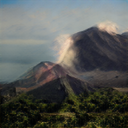} &
\includegraphics[width=0.064\linewidth]{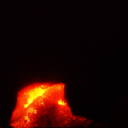} \\
\includegraphics[width=0.064\linewidth]{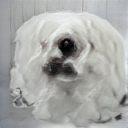} &
\includegraphics[width=0.064\linewidth]{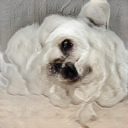} &
\includegraphics[width=0.064\linewidth]{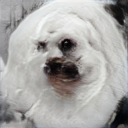} &
\includegraphics[width=0.064\linewidth]{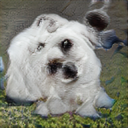} &
\includegraphics[width=0.064\linewidth]{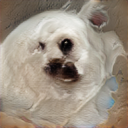} &
&
\includegraphics[width=0.064\linewidth]{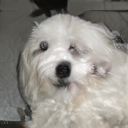} &
\includegraphics[width=0.064\linewidth]{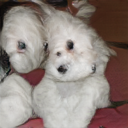} &
\includegraphics[width=0.064\linewidth]{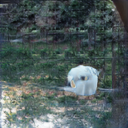} &
\includegraphics[width=0.064\linewidth]{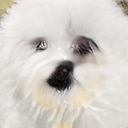} &
\includegraphics[width=0.064\linewidth]{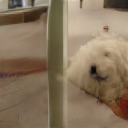} &
&
\includegraphics[width=0.064\linewidth]{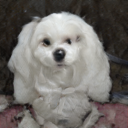} &
\includegraphics[width=0.064\linewidth]{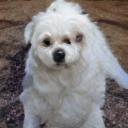} &
\includegraphics[width=0.064\linewidth]{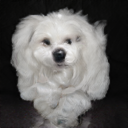} &
\includegraphics[width=0.064\linewidth]{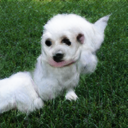} &
\includegraphics[width=0.064\linewidth]{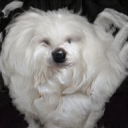} \\
\includegraphics[width=0.064\linewidth]{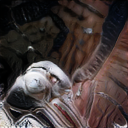} &
\includegraphics[width=0.064\linewidth]{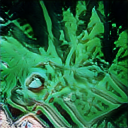} &
\includegraphics[width=0.064\linewidth]{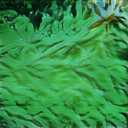} &
\includegraphics[width=0.064\linewidth]{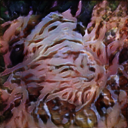} &
\includegraphics[width=0.064\linewidth]{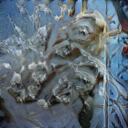} &
&
\includegraphics[width=0.064\linewidth]{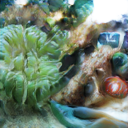} &
\includegraphics[width=0.064\linewidth]{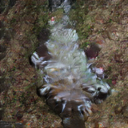} &
\includegraphics[width=0.064\linewidth]{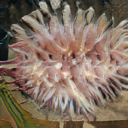} &
\includegraphics[width=0.064\linewidth]{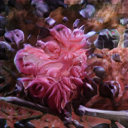} &
\includegraphics[width=0.064\linewidth]{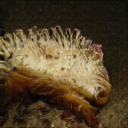} &
&
\includegraphics[width=0.064\linewidth]{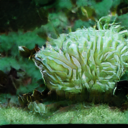} &
\includegraphics[width=0.064\linewidth]{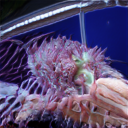} &
\includegraphics[width=0.064\linewidth]{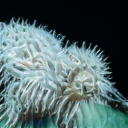} &
\includegraphics[width=0.064\linewidth]{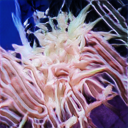} &
\includegraphics[width=0.064\linewidth]{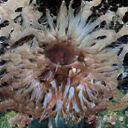} \\
    \end{tabular}\egroup
    \caption{Visual comparison of class-conditional image synthesis results on ImageNet.
      Our method constantly produces plausible images while respecting the given condition.
    }\label{fig:gen_imagenet200_10_05_010} 
\end{figure*}

\begin{figure*}[tb]
\centering
\includegraphics[width=1\linewidth]{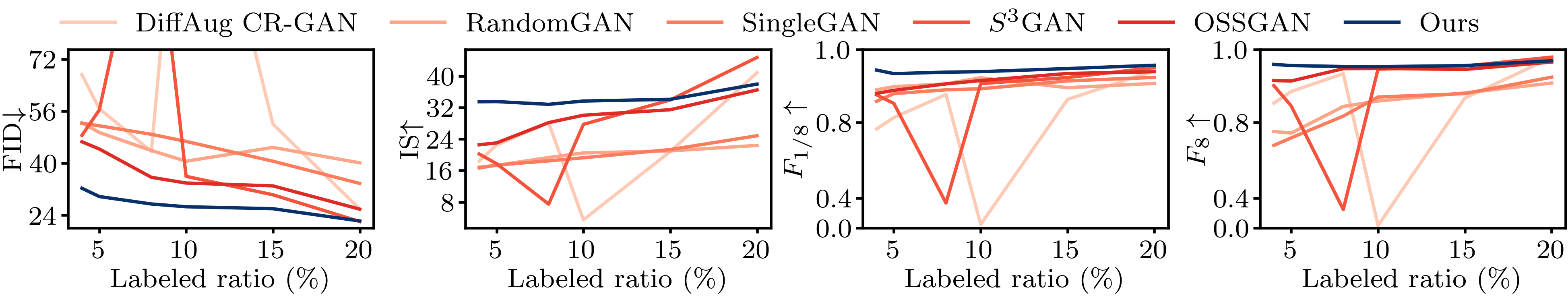}
\caption{
Quantitative comparison over different numbers of labeled samples. We report the results of 
the experiments on the ImageNet dataset with 200 classes, 10\% label noise ratio, and labeled sample ratio of $\{4\%,5\%,8\%,10\%,15\%,20\%\}$.
 Our method outperforms baselines in difficult datasets (blue line).
}\label{fig:numlabeleddata}
\end{figure*}

\begin{table}[tb]
\centering
\caption{Quantitative comparison on ImageNet with closed-set 200 classes, 5\% labeled data, 10\% label-noise.
}\label{tb:imagenet200_10_05_010} 
\resizebox{1\columnwidth}{!}{\begin{tabular}{lccccc}\toprule
& $F_{8}\uparrow$ & $F_{1/8}\uparrow$ & FID$\downarrow$ & IS$\uparrow$ & iFID$\downarrow$ \\\midrule
\baseline & 0.8962 & 0.8171 & 56.7504 & 22.4951  & \sbest{228.9962} \\
RandomGAN & 0.7620 & \sbest{0.9095} & 49.4013 & 17.3114 & 266.7480 \\
SingleGAN & 0.7434 & 0.8903 & 51.4632 & 17.3922 & 292.1951 \\
$S^3$GAN & 0.4078 & 0.5097 & 111.2998 & 8.7617 & 246.3401 \\
OSSGAN & \sbest{0.9245} & 0.8995 & \sbest{44.3262} & \sbest{23.0263} & 238.2692 \\
\ours & \best{0.9630} & \best{0.9433} & \best{29.6751} & \best{33.5418} & \best{183.1367} \\ \bottomrule
\end{tabular}}
\end{table}

\begin{table}[tb]
\centering
\caption{Quantitative comparison on ImageNet $256\times 256$. %
}\label{tb:imagenet256}
\resizebox{1\columnwidth}{!}{\begin{tabular}{lccccc}\toprule
& $F_{8}\uparrow$ & $F_{1/8}\uparrow$ & FID$\downarrow$ & IS$\uparrow$ & iFID$\downarrow$ \\\midrule
\baseline & 0.8177 & 0.7290 & 83.6051 & 20.8947 & 274.5373\\
RandomGAN & 0.7707 & 0.8242 & 60.1051 & 19.3663& 282.4955\\
SingleGAN & 0.8052 & 0.7944 & 62.4891 & 19.4504& 280.3701\\
$S^3$GAN & 0.9002 & \sbest{0.8473} & \sbest{52.2834} & 27.6553 & \sbest{225.6078}\\
OSSGAN & \best{0.9146} & 0.8124 & 53.7868 & \sbest{28.3792} & 229.9876\\
\ours & \sbest{0.9076} & \best{0.8833} & \best{44.5838} & \best{30.0695} & \best{214.7384}\\\bottomrule
\end{tabular}}
\end{table}

\begin{table}[tb]
\centering
\caption{Quantitative comparison of other cGAN models
on TinyImageNet. %
In addition to a projection-based GAN 
, our method shows the performance gain over classifier-based cGAN models.
}\label{tb:adc_tac} 
\resizebox{0.96\columnwidth}{!}{\begin{tabular}{lcccc}\toprule
& \multicolumn{2}{c}{ADC-GAN~\cite{hou2022conditional}}  & \multicolumn{2}{c}{TAC-GAN~\cite{gong2019twin}} \\ \cmidrule(lr){2-3} \cmidrule(lr){4-5}
& FID$\downarrow$ & IS$\uparrow$ & FID$\downarrow$ & IS$\uparrow$  \\\midrule
Supervised & \phantom{0}66.5229 & \phantom{0}8.6387 & 50.4258 &\phantom{0}9.2594 \\
RandomGAN & \phantom{0}40.2519 & 10.5410 & 37.7453 & 10.9988 \\
SingleGAN & \phantom{0}43.6353 & 10.2666 & 38.5622 &10.6992 \\
$S^3$GAN & \phantom{0}50.7904 & 10.0583 & 39.2887 &10.5139 \\
OSSGAN & 113.1070 & \phantom{0}4.7492 & 41.7552 &10.3462 \\
\ours &  \phantom{0}\best{37.0131} & \best{12.1424} & \best{37.4393} & \best{11.3654} \\ \bottomrule
\end{tabular}}
\end{table}

\begin{table}[tb]
\centering
\caption{Quantitative comparison on WebVision~\cite{li2017webvision}.
}\label{tb:webvision_200_2}
\resizebox{1\columnwidth}{!}{\begin{tabular}{lccccc}\toprule
& $F_{8}\uparrow$ & $F_{1/8}\uparrow$ & FID$\downarrow$ & IS$\uparrow$ & iFID$\downarrow$\\\midrule
\baseline & 0.7812 & 0.7725 & 74.3157 & 14.3693  & 249.0955\\
RandomGAN & 0.7840 & 0.8627 & 54.8598 & 14.7182 & 246.9653\\
SingleGAN & 0.7065 & 0.8276 & 64.8178 & 13.5105 & 280.1292\\
$S^3$GAN & \sbest{0.8209} & \sbest{0.8680} & \sbest{63.4304} & 14.6397  & \sbest{238.7989}\\
OSSGAN & 0.7911 & 0.8294 & 66.7111 & \sbest{14.9287} & 242.6553\\
\ours & \best{0.8465} & \best{0.8866} & \best{51.1604} & \best{18.0428} & \best{213.5669} \\ \bottomrule
\end{tabular}}
\end{table}

\noindent \textbf{Evaluation metrics.}
We use
IS~\cite{Salimans2016}, 
FID~\cite{Heusel2017}, iFID, $F_{1/8}$ score~\cite{Sajjadi2018}, and $F_8$ score~\cite{Sajjadi2018}.
FID measures the distance between the generated
and reference images in the feature space using overall data and iFID uses per-class data, but it was not possible to separate
the evaluated values into fidelity and diversity. On the contrary,
$F_{1/8}$ and $F_8$ quantify the fidelity and diversity, respectively.
We sample 10K generated images for all metrics and use the
evaluation set as the reference distribution for FID, iFID, $F_{1/8}$, and $F_8$.

\noindent \textbf{Comprehensive study.}
We first conduct a quantitative study on the TinyImageNet dataset with 150 closed-set classes, 50 open-set classes, 20\% labeled data, and 10\% label noise. Namely, the dataset consists of 15K labeled samples and 85K unlabeled samples.
\Cref{tb:tiny150_10_20} reports the average and standard deviation of FID, IS, $F_{1/8}$, $F_8$, and iFID over three trials.
Our method achieves the best scores in terms of all metrics and achieves tight standard deviations, showing the consistent improvement over the baselines.
On the contrary, the improvement by rcGAN is not the case. In relabeling baselines, only classifier-based GANs improve the performance from naive baselines, because reRandomGAN and reSingleGAN add extra noise strongly.

We then investigate the robustness of the method to label noise in experiments with different label noise ratios.
We show the performance of the methods on different label noise ratios, $\{10\%, 30\%, 50\%, 70\%, 90\%\}$.
Our method still outperforms compared methods even when the labels are considerably noisy (\eg, 90\%), as shown in \cref{fig:labelnoise}. CurriculumGAN easily fails in the experiments in difficult datasets (\eg, 70\% or 90\%).

\noindent \textbf{Ablation study.}
 To evaluate the individual contribution of each component, we carried out an ablation
 study of our method. For this evaluation, we prepare three ablation models: AB1 AB2, and AB3.
 AB1 is equipped with cross entropy loss instead of generalized cross entropy, having lost the robustness to label noise.
 AB2 does not use curriculum learning, assigning equal weights to all samples. The method corrects wrong labels, assigns new labels, and distributes equal weights to all samples, and their classifier is trained on 
generalized cross entropy.
AB3 does not correct the labels of the labeled data. The method assigns new labels to unlabeled data and distributes weights according to the classifier's confidences. It is close to ordinal curriculum learning.
The results of the ablation study on two configurations are given in \cref{tb:ablation}.
With cross entropy, AB1 drops performance, showing the contribution of the robust classifier. 
Since correcting labels of labeled samples without soft curriculum may add extra label noises, AB2 records the worst performance in terms of FID, IS, and iFID in datasets with a large label noise ratio.
AB3 shows a large degradation in the performance under highly noisy data by maintaining label noise.
In both trials, the final model (\ours) enhances the performance of the ablation models by the combination of
robust training and soft curriculum learning.

\noindent \textbf{Evaluation on large datasets.}
We evaluate the proposed method on more complex and challenging datasets to see its stability.
\Cref{tb:imagenet100_10_05_010} show the quantitative results of the ImageNet experiments.
In the experiments, we observe the performance gains over baselines in terms of quantitative metrics.
\Cref{fig:gen_imagenet200_10_05_010,fig:gen_imagenet200_10_05_010_2,tb:imagenet200_10_05_010} show the experimental results on the ImageNet dataset with 200 closed-set classes, 5\% labeled data, 10\% label noise, and 10\% usage ratio. Namely, the dataset has about 12K labeled samples and 345K unlabeled samples.
Our method outperforms all baselines with the quantitative metrics as shown in \cref{tb:imagenet200_10_05_010}.
\Cref{fig:gen_imagenet200_10_05_010_2} demonstrates the fidelity of the images generated by our method.
\Cref{fig:gen_imagenet200_10_05_010} shows in consistency with \cref{tb:imagenet200_10_05_010} that our method
generates images with high fidelity and diversity.
With our soft curriculum, we observe the performance gain over baselines on difficult datasets with limited labeled samples, as shown in \cref{fig:numlabeleddata}. In particular, the proposed approach achieves a competitive performance to semi-supervised and supervised cGANs with 1/3 of the labeled data in terms of FID and IS (5\% vs. 15\%) and half of the labeled data in terms of $F_{1/8}$ and $F_8$ (5\% vs. 10\%).

To demonstrate the effectiveness of our method on high resolution, we conduct experiments on ImageNet $256\times 256$ with 200 closed-set classes, 4\% labeled samples, 10\% label-noise, and 10\% usage ratio.
\Cref{tb:imagenet256} shows that the proposed method outperforms the baselines stably.

\noindent \textbf{Evaluation on classifier-based cGANs.}
Next, we evaluate our method on different cGAN models. In the above evaluations, we build the compared method by integrating semi-supervised methods into projection-based cGANs.
To evaluate the applicability of our method to other cGAN models, we conduct experiments on additional base architectures of classifier-based cGANs (\ie, ADC-GAN~\cite{hou2022conditional} and TAC-GAN~\cite{gong2019twin}).
\Cref{tb:adc_tac} shows our method outperforms baselines in the ADC-GAN and TAC-GAN experiments.

\noindent \textbf{Evaluation on real-world noise.}
Finally, we test our method on WebVision~\cite{li2017webvision} to assess the effectiveness on real-world noise.
WebVision is a dataset built via web queries, and so it contains real-world noise. We use 200 classes as the closed-set classes, drop 98\% labels from the closed-set class samples to make unlabeled data, and the usage ratio of 10\%. \Cref{tb:webvision_200_2} shows the results of the experiments on WebVision. We improve \baseline and achieve an FID of $51.1604$ with an IS of $18.0428$ on the dataset with
real-world noise.

\section{Conclusion}
We presented a novel image generation training framework that allows the training dataset to be composed of noisy labeled and uncurated unlabeled data.
We proposed soft curriculum learning for this new data setting that provides clean labeled data to the discriminator while eliminating the effects of useless samples by correcting noisy labels and assigning new labels.
Concurrently, we use soft labels and generalized cross entropy loss to deal with open-set samples, avoiding overconfidence in samples that do not belong to known classes.
Our comprehensive experiments show that, even when the number of labeled samples is limited and noisy, the proposed method consistently outperforms baselines in both qualitative and quantitative evaluations.
Our method reduces the amount of labeled data required to achieve equivalent performance in the training of conditional GANs.
Furthermore, when tested with different GANs architectures, our method demonstrates stable performance.
We believe that our proposed method expands the real-world applications of cGANs in a sustainable way by making it easier to create datasets for training cGANs.

\noindent \textbf{Limitation}.
Although our method improves baselines on challenge datasets, the beneficial improvement on datasets with sufficient labeled samples is not observed. A deep analysis of the relationship between labeled data size and cGAN performance will provide further insight into the use of our soft curriculum method.

{\small
\bibliographystyle{ieee_fullname}
\bibliography{references}

\begin{thebibliography}{10}\itemsep=-1pt

\bibitem{albert2022addressing}
Paul Albert, Diego Ortego, Eric Arazo, Noel~E O'Connor, and Kevin McGuinness.
\newblock Addressing out-of-distribution label noise in webly-labelled data.
\newblock In {\em WACV}, pages 392--401, 2022.

\bibitem{angluin1988learning}
Dana Angluin and Philip Laird.
\newblock Learning from noisy examples.
\newblock {\em Machine Learning}, 2(4):343--370, 1988.

\bibitem{Brock2018}
Andrew Brock, Jeff Donahue, and Karen Simonyan.
\newblock Large scale {GAN} training for high fidelity natural image synthesis.
\newblock In {\em ICLR}, 2018.

\bibitem{cascante2021curriculum}
Paola Cascante-Bonilla, Fuwen Tan, Yanjun Qi, and Vicente Ordonez.
\newblock Curriculum labeling: Revisiting pseudo-labeling for semi-supervised
  learning.
\newblock In {\em AAAI}, volume~35, 2021.

\bibitem{Ting2019}
Ting Chen, Xiaohua Zhai, Marvin Ritter, Mario Lucic, and Neil Houlsby.
\newblock Self-supervised {GAN}s via auxiliary rotation loss.
\newblock In {\em CVPR}, pages 12146--12155, 2019.

\bibitem{ghosh2017robust}
Aritra Ghosh, Himanshu Kumar, and P~Shanti Sastry.
\newblock Robust loss functions under label noise for deep neural networks.
\newblock In {\em AAAI}, 2017.

\bibitem{gong7465792}
Chen Gong, Dacheng Tao, Stephen~J. Maybank, Wei Liu, Guoliang Kang, and Jie
  Yang.
\newblock Multi-modal curriculum learning for semi-supervised image
  classification.
\newblock {\em IEEE TIP}, 25(7):3249--3260, 2016.

\bibitem{gong2019twin}
Mingming Gong, Yanwu Xu, Chunyuan Li, Kun Zhang, and Kayhan Batmanghelich.
\newblock Twin auxilary classifiers {GAN}.
\newblock {\em NeurIPS}, 32, 2019.

\bibitem{hataya2019unifying}
Ryuichiro Hataya and Hideki Nakayama.
\newblock Unifying semi-supervised and robust learning by mixup.
\newblock In {\em ICLR The 2nd Learning from Limited Labeled Data (LLD)
  Workshop}, 2019.

\bibitem{Heusel2017}
Martin Heusel, Hubert Ramsauer, Thomas Unterthiner, Bernhard Nessler, and Sepp
  Hochreiter.
\newblock {GAN}s trained by a two time-scale update rule converge to a local
  nash equilibrium.
\newblock In {\em NeurIPS}, pages 6626--6637, 2017.

\bibitem{hou2022conditional}
Liang Hou, Qi Cao, Huawei Shen, Siyuan Pan, Xiaoshuang Li, and Xueqi Cheng.
\newblock Conditional {GANs} with auxiliary discriminative classifier.
\newblock In {\em ICML}, pages 8888--8902, 2022.

\bibitem{Kaneko2019}
Takuhiro Kaneko, Yoshitaka Ushiku, and Tatsuya Harada.
\newblock Label-noise robust generative adversarial networks.
\newblock In {\em CVPR}, 2019.

\bibitem{katsumata2022ossgan}
Kai Katsumata, Duc~Minh Vo, and Hideki Nakayama.
\newblock {OSSGAN:} open-set semi-supervised image generation.
\newblock In {\em CVPR}, pages 11185--11193, 2022.

\bibitem{le2015tiny}
Ya Le and Xuan Yang.
\newblock Tiny imagenet visual recognition challenge.
\newblock {\em CS 231N}, 7(7):3, 2015.

\bibitem{li2017webvision}
Wen Li, Limin Wang, Wei Li, Eirikur Agustsson, and Luc Van~Gool.
\newblock Webvision database: Visual learning and understanding from web data.
\newblock {\em arXiv preprint arXiv:1708.02862}, 2017.

\bibitem{Lim2017}
Jae~Hyun Lim and Jong~Chul Ye.
\newblock Geometric {GAN}.
\newblock {\em arXiv preprint arXiv:1705.02894}, 2017.

\bibitem{Lucic2019}
Mario Lu{\v{c}}i{\'c}, Michael Tschannen, Marvin Ritter, Xiaohua Zhai, Olivier
  Bachem, and Sylvain Gelly.
\newblock High-fidelity image generation with fewer labels.
\newblock In {\em ICML}, volume~97, pages 4183--4192, 2019.

\bibitem{Luo2021}
Huixiang Luo, Hao Cheng, Yuting Gao, Ke Li, Mengdan Zhang, Fanxu Meng, Xiaowei
  Guo, Feiyue Huang, and Xing Sun.
\newblock On the consistency training for open-set semi-supervised learning.
\newblock {\em arXiv preprint arXiv:2101.08237}, 2021.

\bibitem{Mirza2014}
Mehdi Mirza and Simon Osindero.
\newblock Conditional generative adversarial nets.
\newblock {\em arXiv preprint arXiv:1411.1784}, 2014.

\bibitem{Miyato2018b}
Takeru Miyato and Masanori Koyama.
\newblock c{GAN}s with projection discriminator.
\newblock In {\em ICLR}, 2018.

\bibitem{natarajan2013learning}
Nagarajan Natarajan, Inderjit~S Dhillon, Pradeep~K Ravikumar, and Ambuj Tewari.
\newblock Learning with noisy labels.
\newblock {\em NeurIPS}, 26, 2013.

\bibitem{ni2022manifold}
Yao Ni, Piotr Koniusz, Richard Hartley, and Richard Nock.
\newblock Manifold learning benefits {GANs}.
\newblock In {\em CVPR}, 2022.

\bibitem{rasmus2015semi}
Antti Rasmus, Mathias Berglund, Mikko Honkala, Harri Valpola, and Tapani Raiko.
\newblock Semi-supervised learning with ladder networks.
\newblock {\em NeurIPS}, 2015.

\bibitem{Olga2015}
Olga Russakovsky, Jia Deng, Hao Su, Jonathan Krause, Sanjeev Satheesh, Sean Ma,
  Zhiheng Huang, Andrej Karpathy, Aditya Khosla, Michael Bernstein,
  Alexander~C. Berg, and Li Fei-Fei.
\newblock {ImageNet} large scale visual recognition challenge.

\bibitem{sachdeva2021evidentialmix}
Ragav Sachdeva, Filipe~R Cordeiro, Vasileios Belagiannis, Ian Reid, and Gustavo
  Carneiro.
\newblock {EvidentialMix:} learning with combined open-set and closed-set noisy
  labels.
\newblock In {\em WACV}, pages 3607--3615, 2021.

\bibitem{saito2021openmatch}
Kuniaki Saito, Donghyun Kim, and Kate Saenko.
\newblock {OpenMatch:} open-set semi-supervised learning with open-set
  consistency regularization.
\newblock In {\em NeurIPS}, 2021.

\bibitem{Sajjadi2018}
Mehdi~SM Sajjadi, Olivier Bachem, Mario Lucic, Olivier Bousquet, and Sylvain
  Gelly.
\newblock Assessing generative models via precision and recall.
\newblock In {\em NeurIPS}, pages 5234--5243, 2018.

\bibitem{Salimans2016}
Tim Salimans, Ian Goodfellow, Wojciech Zaremba, Vicki Cheung, Alec Radford, and
  Xi Chen.
\newblock Improved techniques for training {GAN}s.
\newblock {\em NeurIPS}, 29:2234--2242, 2016.

\bibitem{thekumparampil2018robustness}
Kiran~K Thekumparampil, Ashish Khetan, Zinan Lin, and Sewoong Oh.
\newblock Robustness of conditional {GANs} to noisy labels.
\newblock {\em NeurIPS}, 31, 2018.

\bibitem{Tran2017}
Dustin Tran, Rajesh Ranganath, and David Blei.
\newblock Hierarchical implicit models and likelihood-free variational
  inference.
\newblock In {\em NeurIPS}, pages 5523--5533, 2017.

\bibitem{wang2018iterative}
Yisen Wang, Weiyang Liu, Xingjun Ma, James Bailey, Hongyuan Zha, Le Song, and
  Shu-Tao Xia.
\newblock Iterative learning with open-set noisy labels.
\newblock In {\em CVPR}, pages 8688--8696, 2018.

\bibitem{TinyImageNet}
Jiayu Wu, Qixiang Zhang, and Guoxi Xu.
\newblock Tiny {ImageNet} challenge.

\bibitem{yao2021jo}
Yazhou Yao, Zeren Sun, Chuanyi Zhang, Fumin Shen, Qi Wu, Jian Zhang, and
  Zhenmin Tang.
\newblock {Jo-SRC:} a contrastive approach for combating noisy labels.
\newblock In {\em CVPR}, pages 5192--5201, 2021.

\bibitem{Yu2020}
Qing Yu, Daiki Ikami, Go Irie, and Kiyoharu Aizawa.
\newblock Multi-task curriculum framework for open-set semi-supervised
  learning.
\newblock In {\em ECCV}, pages 438--454, 2020.

\bibitem{zhang2021flexmatch}
Bowen Zhang, Yidong Wang, Wenxin Hou, Hao Wu, Jindong Wang, Manabu Okumura, and
  Takahiro Shinozaki.
\newblock Flexmatch: Boosting semi-supervised learning with curriculum pseudo
  labeling.
\newblock {\em NeurIPS}, 34:18408--18419, 2021.

\bibitem{Zhang2019}
Han Zhang, Ian Goodfellow, Dimitris Metaxas, and Augustus Odena.
\newblock Self-attention generative adversarial networks.
\newblock In {\em ICML}, pages 7354--7363, 2019.

\bibitem{zhang2019consistency}
Han Zhang, Zizhao Zhang, Augustus Odena, and Honglak Lee.
\newblock Consistency regularization for generative adversarial networks.
\newblock In {\em ICLR}, 2019.

\bibitem{zhang2018generalized}
Zhilu Zhang and Mert Sabuncu.
\newblock Generalized cross entropy loss for training deep neural networks with
  noisy labels.
\newblock {\em NeurIPS}, 31, 2018.

\bibitem{Zhao2020}
Shengyu Zhao, Zhijian Liu, Ji Lin, Jun-Yan Zhu, and Song Han.
\newblock Differentiable augmentation for data-efficient {GAN} training.
\newblock In {\em NeurIPS}, 2020.

\end{thebibliography}
}

\end{document}